RESEARCH ARTICLE

# Enabling Large Batch Size Training for DNN Models Beyond the Memory Limit While Maintaining Performance

**XINYU PIAO [1], DOANGJOO SYNN [2], JOOYOUNG PARK [3], AND JONG-KOOK KIM [1], (Senior Member, IEEE)**

[1] School of Electrical Engineering, Korea University, Seoul 02841, South Korea
[2] School of Computer Science, Georgia Institute of Technology, Atlanta, GA 30332, USA
[3] School of Computer Science and Engineering, Nanyang Technological University, Singapore 639798

Corresponding author: Jong-Kook Kim (jongkook@korea.ac.kr)

This work was supported by the Basic Science Research Program through the National Research Foundation of Korea (NRF) funded by the Ministry of Education under Grant NRF-2016R1D1A1B04933156.

**ABSTRACT** Recent deep learning models are difficult to train using a large batch size, because commodity machines may not have enough memory to accommodate both the model and a large data batch size. The batch size is one of the hyper-parameters used in the training model, and it is dependent on and is limited by the target machine memory capacity because the batch size can only fit into the remaining memory after the model is uploaded. Moreover, the data item size is also an important factor because if each data item size is larger then the batch size that can fit into the remaining memory becomes smaller. This paper proposes a method called Micro-Batch Processing (MBP) to address this problem. This method helps deep learning models to train by providing a batch processing method that splits a batch into a size that can fit in the remaining memory and processes them sequentially. After processing the small batches individually, a loss normalization algorithm based on the gradient accumulation is used to maintain the performance. The purpose of our method is to allow deep learning models to train using larger batch sizes that exceed the memory capacity of a system without increasing the memory size or using multiple devices (GPUs).

## I. INTRODUCTION

Recently, many researches use heavy Deep Neural Network (DNN) models that need a lot of memory to execute. Plus, data sizes are increasing such that it is difficult to increase the mini-batch size which is used for the training of the models. The mini-batch size is an important hyper-parameter that determines the number of data set items that are used in one iteration of a training process and the mini-batch size may affect the overall performance of the DNN ([1], [2]). The results in table 1 show the comparison between a large mini-batch and a small mini-batch used for the image classification and semantic segmentation problems. Utilizing a large mini-batch shows 21.88% anxd 2.01% higher than a small mini-batch when using higher resolution images for image classification and semantic segmentation problems, respectively.

The image size may also affect the model performance, because the image size may limit the mini-batch size. Also, higher resolution images (i.e., larger image size) contain more information about objects. Table 1 shows the comparison of results between the higher resolution images and the lower resolution images that are used for the image classification and semantic segmentation problems. Using the higher resolution image data shows 21.64% and 2.01% higher than the lower resolution image data in a large mini-batch for image classification and semantic segmentation problems, respectively. This aspect is also similar when using a small mini-batch size.

The total memory size of a mini-batch can only be increased to the remaining memory after the model is loaded. Thus, there is a limit on the size of the mini-batch and the number of data items included in the mini-batch. If the memory requirement of a certain mini-batch size is larger than the free remaining memory size, the mini-batch cannot be allocated to the GPU memory and the model cannot be trained. If the optimal mini-batch size for a particular resolution dataset is larger than the device memory, then the mini-batch size must be reduced when increasing the image resolution or use low-resolution image data to increase the mini-batch size. Therefore, the significant increase in image size makes it more challenging to train DNN models.

**TABLE 1. The effect of batch size and image size for ResNet-50 model using Flower-102 dataset and U-Net model using Carvana dataset.**

| Model | | ResNet-50 | | U-Net | |
|---|---|---|---|---|---|
| Metric | | Max. acc (%) | | Max. IoU (%) | |
| Image size | | 32×32 | 224×224 | 96×96 | 384×384 |
| Batch Size | 2 | 48.66 | 61.86 | 92.30 | 93.61 |
| | 16 | 62.10 | **83.74** | 93.61 | **95.62** |

Many researchers tried various parallelism techniques to alleviate the problems that deep learning methods face as shown in figure 1a∼1c. Data parallelism (e.g., [3], [4], [5]) is usually used when the mini-batch size is too large to be fit into a single device's memory (shown in figure 1a). A mini-batch is partitioned for computation and distributed to multiple devices and each device has a full copy of the learning model. When all data within a mini-batch is processed, gradients and weights are updated across devices through communications. Model parallelism partitions the learning model into cells and distributes the cells to multiple devices (figure 1b). It is usually used when the model is too large to fit into a device's memory (e.g., [6], [7], [8]). A mini-batch is initially allocated to the first device that has the first learning layer and then the outcome is set to the successor device via communications until the last device that has the last layer of the model. Research papers that use pipeline parallelism are proposed (e.g., [9], [10]) and it is shown in figure 1c. Pipeline parallelism also partitions the learning model across multiple device workers, such as model parallelism, and maximizes device utilization by splitting a mini-batch into small batches and processing them in a pipelined fashion. These methods employ Data-Parallel Synchronous Stochastic Gradient Descent (SGD). Although, all of these research improve the learning and performance of the models, they still have the problem of the mini-batch size being limited by the device memory size.

This paper proposes Micro-Batch Processing (MBP) that can utilize a large batch of data to train models even if the batch cannot fit into the memory without increasing the device memory or using multiple devices (shown in figure 2a). Thus, allowing researchers to experiment using large mini-batch sizes on a single device without any complex techniques for synchronizations and communications, and without requiring large memory systems. The idea is to split a mini-batch into $n$ micro-batches (e.g., [9], [10]) and stream them sequentially to a GPU, which sequentially computes and trains the learning model based on micro-batches. Results show that MBP can increase the training batch size to the full size of the training set regardless of the model type, dataset, and data size. To maintain the performance, MBP computes the gradient for a large batch using loss normalization based on micro-batch gradient accumulation, a method that accumulates all of the individual gradient calculated for each of the multiple micro-batches.

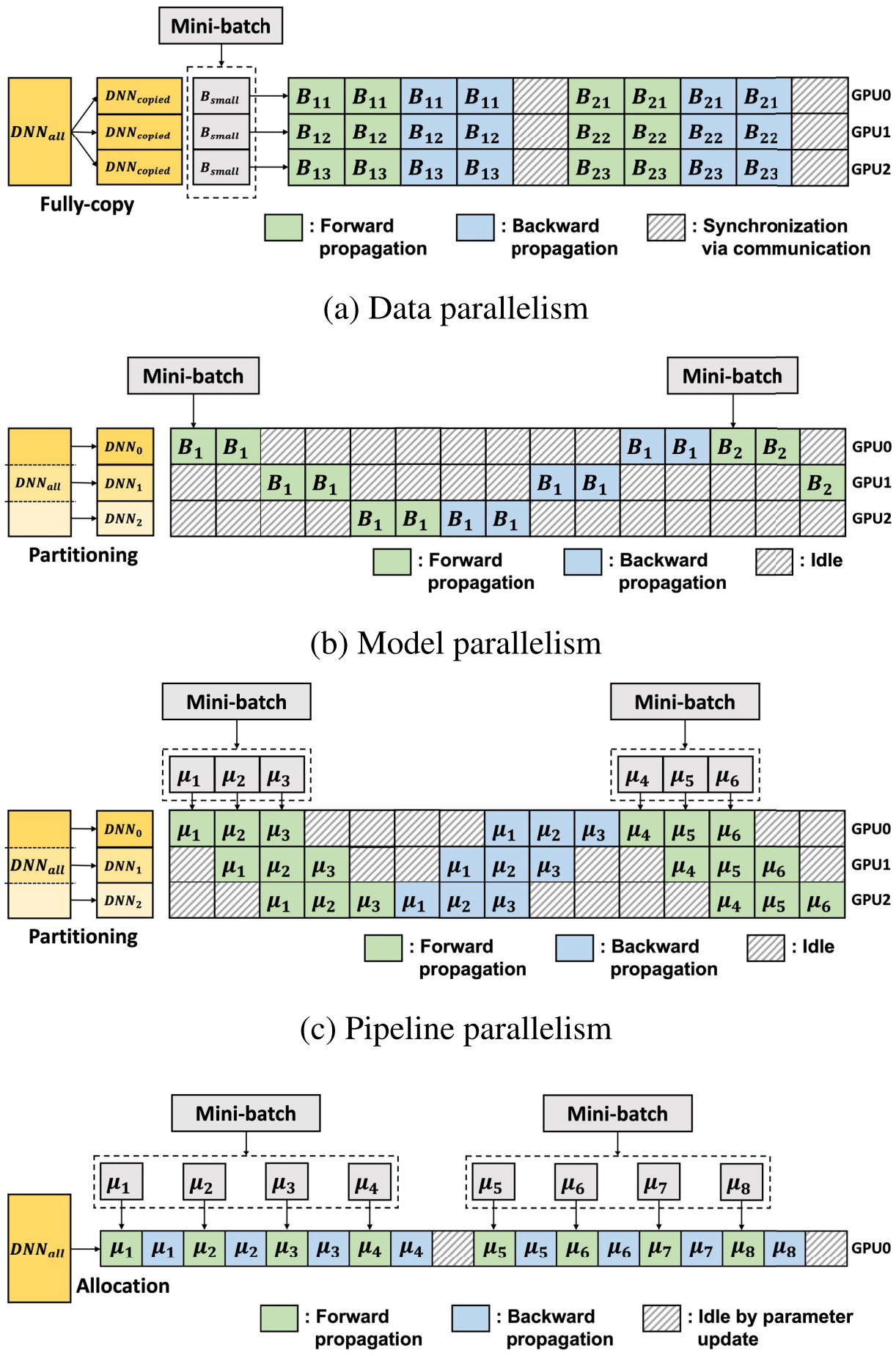

(a) Data parallelism

(b) Model parallelism

(c) Pipeline parallelism

(d) Our proposed method

**FIGURE 1. Difference in the training process between our proposed method and other parallelism approaches.**

This paper is organized as follows. Related work is described in Section II. Section III introduces MBP, terminologies used throughout this paper is described, and the loss normalization algorithm used to maintain the performance is introduced. Section IV shows the results of classification and semantic segmentation tasks. Finally, Section V concludes the paper.

## II. RELATED WORK

If model parallelism is done naively, GPUs suffer from idle time overhead because a worker have to wait for previous or successor worker to finish its job. To minimize this overhead, GPipe [9], a pipeline parallelism library for training large neural networks, presented a novel pipeline parallelism that uses batch splitting. GPipe splits mini-batch into smaller batches, called micro-batch, for pipeline execution across multiple GPUs. This enables each GPU to work on different micro-batch simultaneously, and gradient and model parameters are updated synchronously after all micro-batches are processed (figure 1c). However, synchronizing at the end of every mini-batch still causes GPU to stall. Pipedream [10] is also a pipeline parallelism method that uses asynchronous communication and corresponding work scheduling and learning method to handle this problem.

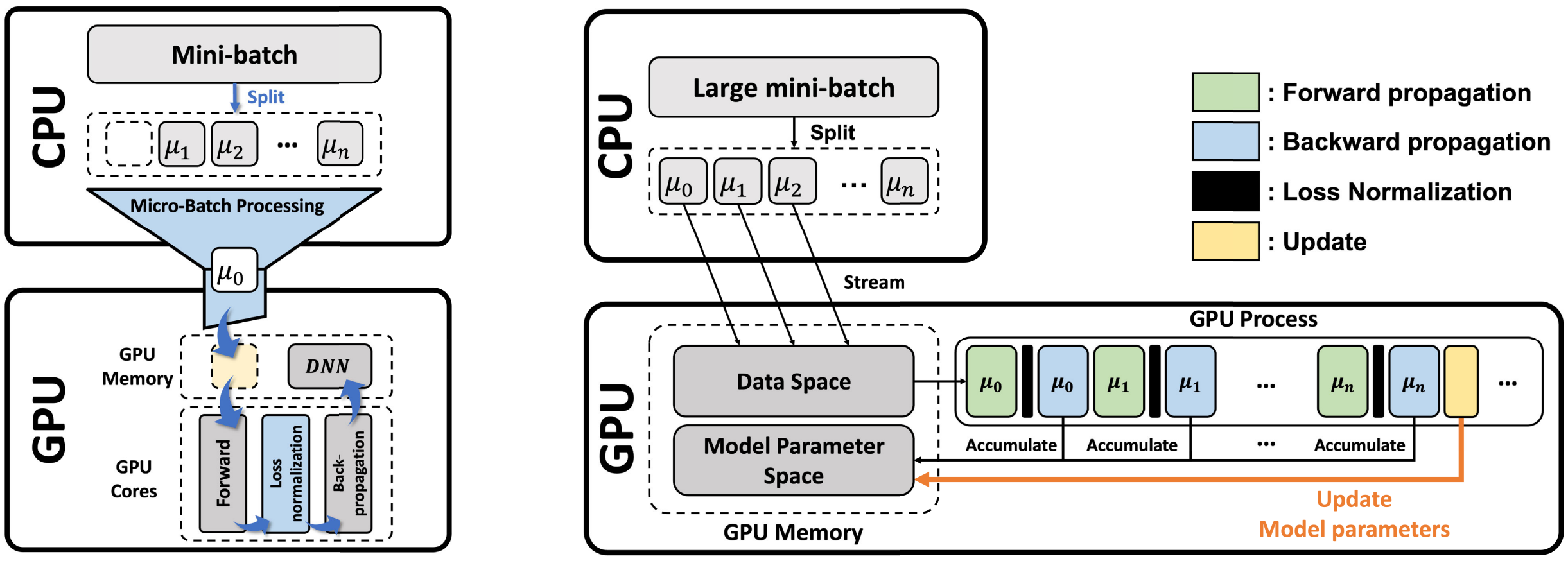

(a) Design of MBP. (b) A system perspective of MBP.

**FIGURE 2. Overview of Micro-Batch Processing (MBP).**

Previous solutions are not designed specifically for large mini-batch training and they focus on maximizing the learning performance or parallelization of the learning process. They also have a limitation where the batch size cannot exceed the remaining memory space after the model is uploaded as the physical memory does not have infinite capacity. This is because previous approaches are typically based on a scheme that stores a mini-batch to be in the GPU memory before starting to train. Thus, training fails if the mini-batch cannot be uploaded to the GPU memory. The proposed method in this paper solves this limitation without increasing the memory of a single device and without using multiple devices (GPUs).

## III. PROPOSED APPROACH

### A. OVERVIEW

Typically, the batch data is pre-loaded in the main memory before being sent and trained in the GPU. Because of the limited size of the GPU's memory, it may not be possible to load a desired batch size of data onto the GPU memory for training. Usually, the CPU sends the pre-loaded batch to the GPU without considering the remaining GPU memory size. Thus, if the remaining GPU memory is not enough to store the batch, the learning model may fail to train.

Micro-Batch Processing (MBP) method enables the learning model to train large mini-batches in a single GPU without increasing the GPU memory and without using multiple GPUs. The key idea of MBP is to split a mini-batch into smaller micro batches and stream them sequentially to a GPU, where each micro batch will be sequentially processed and then the results are accumulated such that it is equal to using a mini-batch that exceed the remaining memory used to train the DNN model.

### B. MICRO-BATCH PROCESSING

The proposed MBP can be located in between the CPU and the GPU to transfer the batch that is needed for the training process and it works independently of the training and the data loading (figure 2a). The MBP uses two technical methods to train large mini-batch; one is a batch streaming method, and the other is the gradient accumulation method. The former method is to split a mini-batch into smaller batches and stream them to the remaining GPU memory, sequentially. Micro-batches can be determined such that it can fit in the remaining GPU memory capacity. MBP also uses the gradient accumulation method that accumulates all of the gradient information of each of the micro-batches. Using small batches to train usually results in poor performance. The gradient accumulation method maintains the training performance as if using a mini-batch that includes multiple micro-batches.

Therefore, the MBP enables using a large mini-batch to train a DNN model that exceed the remaining memory space after the loading of the learning model because MBP trains micro batches and accumulates the gradient information for the mini-batch.

### C. MICRO-BATCH

In MBP, the micro-batch ($\mu$-batch) in [9] is extended and is defined as; 1) a unit of streaming to the target device, and 2) a unit of computing on the target device. The first definition presents the splitting of the input mini-batch into micro batches that can fit onto the remaining memory space and streaming them sequentially to the target device. The second statement describes the unit size that target device uses to train the neural network. The formulation of the micro-batch can be expressed as follows.

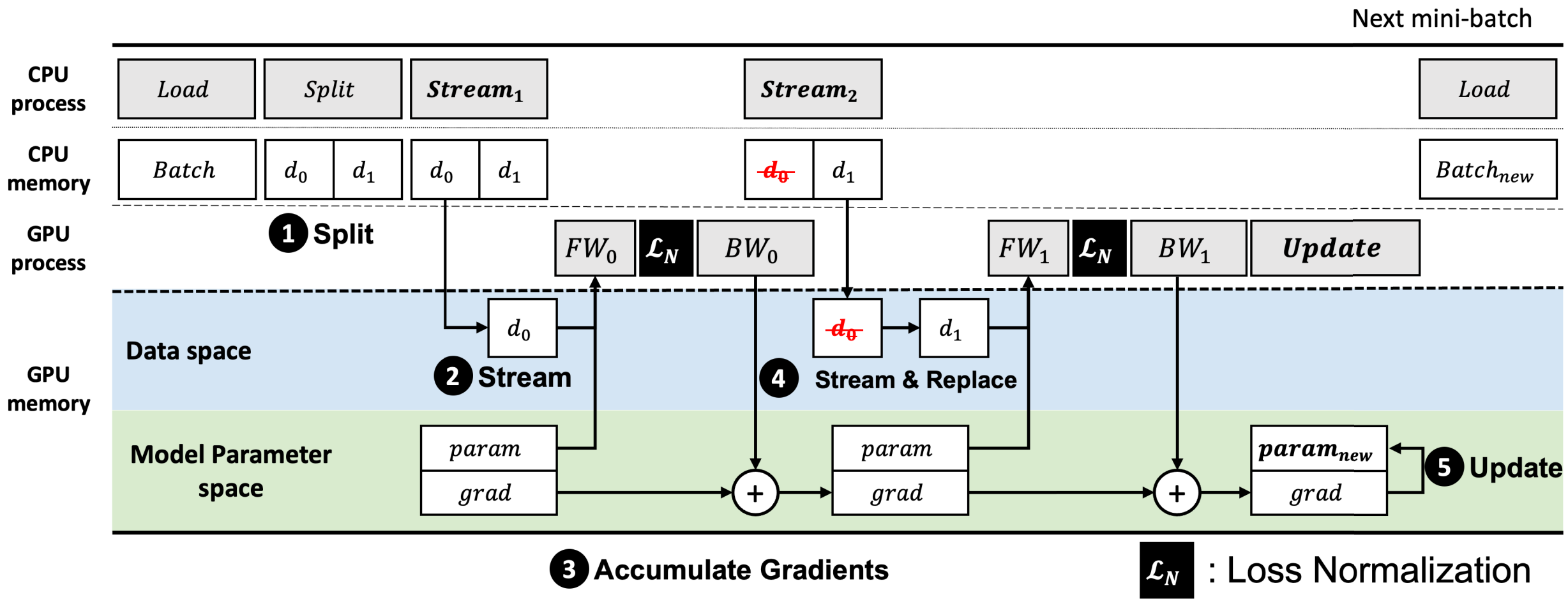

**FIGURE 3.** The training process using MBP.

$$\text{mini-batch} = \bigcup_{i \in S_\mu} \mu\text{-batch}_i \tag{1}$$

$$\mu\text{-batch}_i \subset \text{mini-batch } (i \in S_\mu) \tag{2}$$

$$\mu\text{-batch}_{\text{size}} \leq \text{mini-batch}_{\text{size}} \tag{3}$$

where $S_\mu$ in equation 1 and 2 is the set of micro-batches for that certain mini-batch. The micro-batch effectively removes the dependency between the mini-batch size and the target device's memory capacity. While MBP is using micro-batch for training, this information is accrued such that the model can be trained as if using a large mini-batch size.

### D. TRAINING PROCESS USING MBP

Figure 2b and 3 show the details of the training process using MBP. The input mini-batch is split into micro-batches in the CPU, and these micro-batches are sequentially streamed by the CPU to the GPU for computation. When the learning model starts to train, the GPU memory is allocated and split into two domains; one is the model parameter space where model parameters and gradients reside, and the other is the data space where tensorized input batch and intermediate outcomes are populated. The data in the model parameter space is used to determine the updates of the deep learning model's parameters. The intermediate outcomes that are computed by the forward propagation of the model and in the data space are used to calculate the gradients in the following processes.

The training process using the MBP is as follows. First, MBP loads the mini-batch dataset to the CPU memory space and then splits the mini-batch into $n$ micro-batches in preparation to stream to the GPU (❶). Then, the micro-batches are streamed sequentially to the GPU and the GPU starts to train the model using each micro-batch (❷). The GPU executes the forward and backward propagation steps and stores the gradient in the model parameter space (❸). Only when all the micro-batches in the mini-batch are used for training, the model parameters are updated. The gradients computed by the forward and backward propagation are accumulated until all micro-batch is processed (❹). When the final micro-batch is completed, MBP updates the model parameters using the accumulated gradient, similar to the gradient computed by the mini-batch (❺). Therefore, it is as though the model's parameters are updated using the mini-batch, while the gradients are computed from micro-batches.

The accumulated gradient is computed using the gradient accumulation method that adds the earlier gradients and current gradient until a update. This results in the accumulated gradient being larger than the gradient computed by using a mini-batch. This is because the training process of deep learning model using the MBP and using the mini-batch is different. Thus, the resulting performance can be different. This paper introduces a loss normalization method that normalizes the loss calculated using MBP by a loss function at every micro-batch iteration. The necessity and appropriateness of loss normalization is described in detail in the next section. In MBP, the loss normalization method is used for the automatic update of gradients and MBP waits until all micro-batches, split from a mini-batch, are completed before the update. The timing of the update will seem to be the same as an update after a mini-batch is completed.

### E. LOSS NORMALIZATION

Stochastic Gradient Descent (SGD) [11] is one of the widely used method for deep learning. SGD uses small data set, called mini-batch, randomly selected from the dataset and computes the loss from the difference between outputs computed from the forward pass in the model and targets. MBP adopted the gradient accumulation to derive the gradient of the mini-batch using gradients calculated and accumulated from using micro-batches. If loss accumulation is performed without loss normalization in the micro-batch loss calculation, loss calculated at every micro-batch unit will be accumulated, leading to a different loss calculation compared to the loss calculated without MBP. This section defines the problem of gradient accumulation in the case of training using MBP and proves the necessity and appropriateness of the loss normalization method introduced in this paper.

Gradient accumulation calculates the sum of the previous gradient and the current gradient until updated ($\nabla\omega_{accum} = \sum_{i=0}^{n} \nabla\omega_i$). The formulation of loss ($\mathcal{L}_B$) and gradient ($\nabla\omega_B$) for a mini-batch can be expressed as follows:

$$\mathcal{L}_B = \frac{1}{N_B}\sum_{i=1}^{N_B} \mathcal{L}(\text{output}_i, \text{target}_i) = \frac{1}{N_B}\sum_{i=1}^{N_B} \mathcal{L}_i \quad (4)$$

$$\nabla\omega_B = \partial(\mathcal{L}_B) = \partial(\frac{1}{N_B}\sum_{i=1}^{N_B} \mathcal{L}_i) \quad (5)$$

$$N_B = N_{S_\mu} \times N_\mu \quad (6)$$

where $\mathcal{L}_i$ is loss that is calculated between output and target for the $i$-th batch in one mini-batch, $N_{S_\mu}$ is the size of set of micro-batches in a mini-batch, $N_\mu$ is the size of the micro-batch, and $N_B$ is the size of the mini-batch. The formulation of loss between a mini-batch and micro-batches can be defined as follows (equation 7 and 8):

$$\sum_{i=1}^{N_B} \mathcal{L}_i = \sum_{j=1}^{N_{S_\mu}}\sum_{k=1}^{N_\mu} \mathcal{L}_{j,k} \quad (7)$$

$$\frac{1}{N_B}\sum_{i=1}^{N_B} \mathcal{L}_i = \frac{1}{N_{S_\mu} \cdot N_\mu}\sum_{j=1}^{N_{S_\mu}}\sum_{k=1}^{N_\mu} \mathcal{L}_{j,k} \quad (8)$$

The relationship of the gradient between a mini-batch and micro-batches can be expressed as follows:

$$\nabla\omega_B = \partial(\frac{1}{N_B}\sum_{i=1}^{N_B} \mathcal{L}_i) = \partial(\frac{1}{N_{S_\mu} \cdot N_\mu}\sum_{j=1}^{N_{S_\mu}}\sum_{k=1}^{N_\mu} \mathcal{L}_{j,k}) \quad (9)$$

$$= \partial(\frac{1}{N_{S_\mu}}\sum_{k=1}^{N_{S_\mu}}(\frac{1}{N_\mu}\sum_{j=1}^{N_\mu} \mathcal{L}_j)_k) \quad (10)$$

$$= \partial(\frac{1}{N_{S_\mu}}\sum_{k=1}^{N_{S_\mu}}(\mathcal{L}_\mu)_k) \quad (11)$$

where $\nabla\omega_B$ is the gradient of one mini-batch, and $\mathcal{L}_\mu$ is the loss of one micro-batch. The accumulated gradient ($\nabla\omega_{accum}$) can be expressed as follows:

$$\nabla\omega_{accum} = \sum_{k=1}^{N_{S_\mu}}(\nabla\omega_\mu)_k = \sum_{k=1}^{N_{S_\mu}}\partial(\mathcal{L}_\mu)_k = \partial(\sum_{k=1}^{N_{S_\mu}}(\mathcal{L}_\mu)_k) \quad (12)$$

The above equations (from equation 11 to equation 12) prove that the accumulated gradient is not equal to the gradient of a mini-batch. This means that there is a need for a method to normalize the gradients of a mini-batch that uses the MBP.

**Algorithm 1** Loss Normalization Algorithm

**Require:** The size of micro-batch $N_\mu$, a mini-batch $B$

Count the size of mini-batch $N_B$ from $B$
**if** $N_B < N_\mu$ **then**
  $N_\mu \leftarrow N_B$
**end if**
$N_{S_\mu} \leftarrow$ Round-up($N_B$ / $N_\mu$)
List of micro-batches ($\mu$) $\leftarrow$ Split($B$, $N_\mu$)
$i \leftarrow 0$
**while** $i < N_{S_\mu}$ **do**
  Forward-propagation using micro-batch ($\mu_i$)
  $\mathcal{L}_{\mu_i} \leftarrow \mathcal{L}(\text{output}_i, \text{target}_i)$
  $\mathcal{L}_{\text{norm}_i} \leftarrow \mathcal{L}_{\mu_i}/N_{S_\mu}$
  Backward-propagation and accumulate gradients using $\mathcal{L}_{\text{norm}_i}$
  $i \leftarrow i + 1$
**end while**
Update the DNN model parameters using accumulated gradients by optimizer

To calculate the gradient of mini-batch, we considered two normalization methods; one is normalizing the accumulated gradient for each layer, the other is normalizing loss computed by the loss function. These two methods provide the same effect of normalizing the gradient. However, the former method uses a complex algorithm to normalize the accumulated gradients for each layer, and it will cause heavy overhead at run-time. The latter method does not have a heavy overhead at run-time because it is a simple method that normalizes the accumulated loss (equation 13). Therefore, the latter method to normalize the gradient is used for MBP because this method provides the similar effect of normalizing and is very simple.

$$\mathcal{L}_{\text{norm}_j} = \frac{1}{N_{S_\mu}} \times \mathcal{L}_{\mu_j} \quad (13)$$

$$\sum_{j=1}^{N_{S_\mu}}\partial(\mathcal{L}_{\text{norm}_j}) = \sum_{j=1}^{N_{S_\mu}}\partial(\frac{1}{N_{S_\mu}} \times \mathcal{L}_{\mu_j}) \quad (14)$$

$$= \sum_{j=1}^{N_{S_\mu}}\partial(\frac{1}{N_{S_\mu}} \times \frac{1}{N_\mu}\sum_{k=1}^{N_\mu} \mathcal{L}_{j,k}) \quad (15)$$

$$= \partial(\frac{1}{N_{S_\mu} \cdot N_\mu}\sum_{j=1}^{N_{S_\mu}}\sum_{k=1}^{N_\mu} \mathcal{L}_{j,k}) \quad (16)$$

The above equations (from equation 14 to equation 16) prove that the loss normalization provides the same calculation compared to equation 9. The loss normalization algorithm dynamically determines the normalizing factor and normalizes the loss considering the current input mini-batch size because the mini-batch size is not guaranteed to be uniform for all iterations (shown in Algorithm 1).

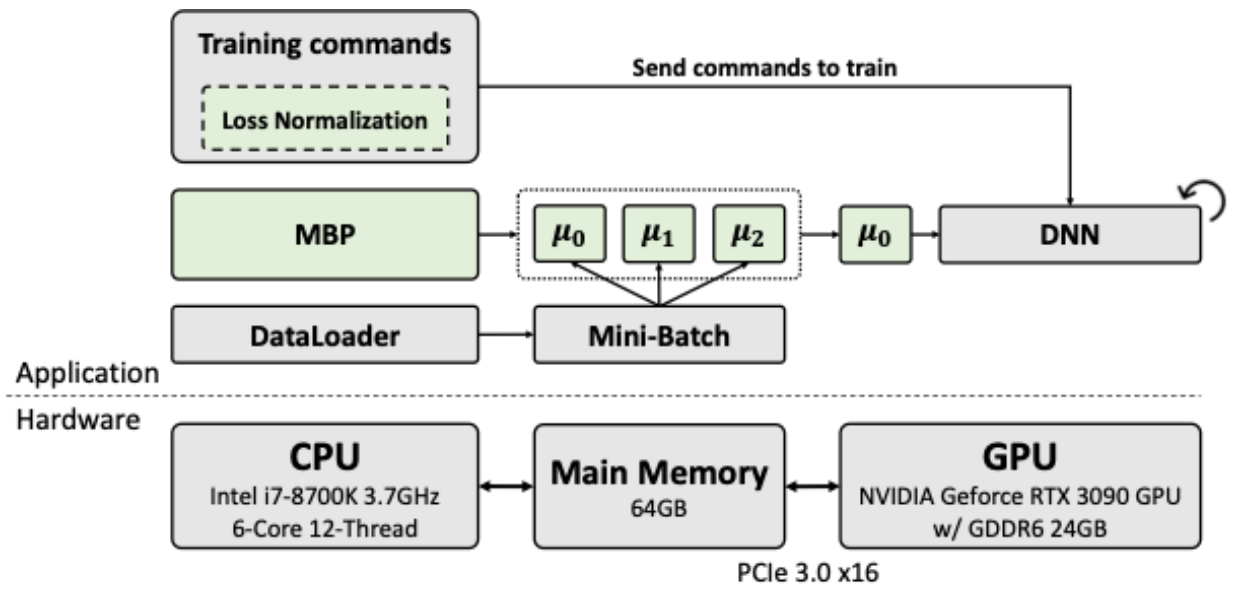

**FIGURE 4.** The diagram of the experimental system setup.

## IV. EVALUATION

### A. EXPERIMENTAL SETUP

#### 1) ENVIRONMENT

As described in section III, MBP enables the training of large mini-batches that do not fit into the remaining memory in a single device without increasing memory size and without using multiple GPUs. Thus, the environment used is a single GPU device system to evaluate the performance. The experiments are run on a system consisting of NVIDIA GeForce RTX 3090 GPU with 24GB GDDR6 memory, intel i7-8700K 3.7GHz 6-Core processor, and 64GB main memory. MBP was implemented by using PyTorch version 1.10.2 and CUDA version 11.3.

#### 2) MODELS

MBP's performance is evaulated using three different types of model architectures. For image classification tasks, two ResNet models [12] (ResNet-50 and ResNet-101)[1] and the AmoebaNet [13] model that uses the architecture search algorithm are used for comparison. Specifically, the AmoebaNet-D model that is used consists of 6 layers and 190 filters and is implemented in [9].[2] For semantic segmentation tasks, the U-Net [14] model,[3] commonly used in many research, is used for testing MBP. The batch normalization [15] layers, which compute the statistics along the input mini-batch dimension, are included in the experiment models.

#### 3) DATASETS

To evaluate the performance of the proposed MBP, two datasets are used that consist of large image size data. This allows for the testing of large batch sizes that exceed the remaining memory capacity. One is the Flower-102 dataset [16] which includes 8,189 color images in 102 classes, where each class consists of between 40 and 258 images, for image classification tasks. The image size of Flower-102 dataset varies, ranging from 500 $\times$ 500 up to 1168 $\times$ 500. Carvana[4] dataset which consists of both color images and binary images of 5,088 1918 $\times$ 1280 image resolution for semantic segmentation tasks.

[1]https://github.com/weiaicunzai/pytorch-cifar100
[2]https://github.com/kakaobrain/torchgpipe
[3]https://github.com/milesial/Pytorch-UNet
[4]https://www.kaggle.com/c/carvana-image-masking-challenge

The original image size is not appropriate for comparison between the learning models that use or not use MBP. Because if the image size is too large, the baseline models that do not use MBP will fail even after 2 mini-batch size because of the memory limitations. Thus, the image size of the given datasets are resized. The image size of Flower-102 dataset was fixed to 224 $\times$ 224, which is widely used image size is other research such as [9], [17], and [12] in image classification tasks, for the ResNet models. For the AmoebaNet-D model, the image size of Flower-102 dataset was fixed at 416 $\times$ 416 image size which is used in [9][2]. The image size of Carvana dataset was fixed at 384 $\times$ 384 as used in [18] using the method of random crop from a resized 959 $\times$ 640 image, which is a method to crop the given image at random location, for U-Net models.

#### 4) METHODS

ResNet-50 and ResNet-101 models are both trained using the same cross-entropy loss function and the SGD optimizer using parameters that are set to 0.01 learning rate, 0.9 momentum, and 0.0005 decay. AmoebaNet-D model is trained using the cross-entropy loss function, SGD optimizer using parameters that are set to 0.1 learning rate, 0.9 momentum, and 0.0001 decay. AmoebaNet-D uses a learning rate scheduler that reduces the learning rate linearly during training.

U-Net model is trained using binary cross-entropy (BCE) loss and dice coefficient (DC), Adam optimizer using parameters that are set to 0.01 learning rate, 0.0005 decay, and without learning rate scheduler. The dice coefficient [19] is a widely used metric to measure the similarity between images for semantic segmentation, and is a value between 0 and 1. The dice coefficient can be expressed as follows:

$$DC = \frac{2|A \cap B|}{|A| + |B|} \tag{17}$$

where $A$ means the ground-truth image, and $B$ means the binary image predicted/computed by the learning model. And then, we use the dice coefficient as loss function and combine it with the binary cross entropy loss. The loss of dice coefficient ($\mathcal{L}_{\text{dc}}$) and the combined loss ($\mathcal{L}_{\text{total}}$) can be expressed as follows:

$$\mathcal{L}_{\text{dc}} = 1 - DC = 1 - \frac{2|A \cap B|}{|A| + |B|} \tag{18}$$

$$\mathcal{L}_{\text{total}} = \mathcal{L}_{\text{bce}} + \mathcal{L}_{\text{dc}} \tag{19}$$

Note that MBP uses the default training algorithm in PyTorch [20] to evaluate performance and pre-trained models are not used in the experiments.

### B. EXPERIMENTAL RESULTS

#### 1) TRAINING PERFORMANCE

The training performance of MBP in all models is shown using the size of mini-batch and micro-batch shown in table 2. The size of mini-batch is determined to be the maximum size that can be computed on the GPU and the size of micro-batch

**TABLE 2. Initial size of mini-batch and micro-batch for different models.**

| Tasks | Model | mini-batch | $\mu$-batch |
|---|---|---|---|
| Classification | ResNet-50 | 16 | 8 |
| | ResNet-101 | 8 | 4 |
| | AmoebaNet-D | 32 | 16 |
| Segmentation | U-Net | 16 | 8 |

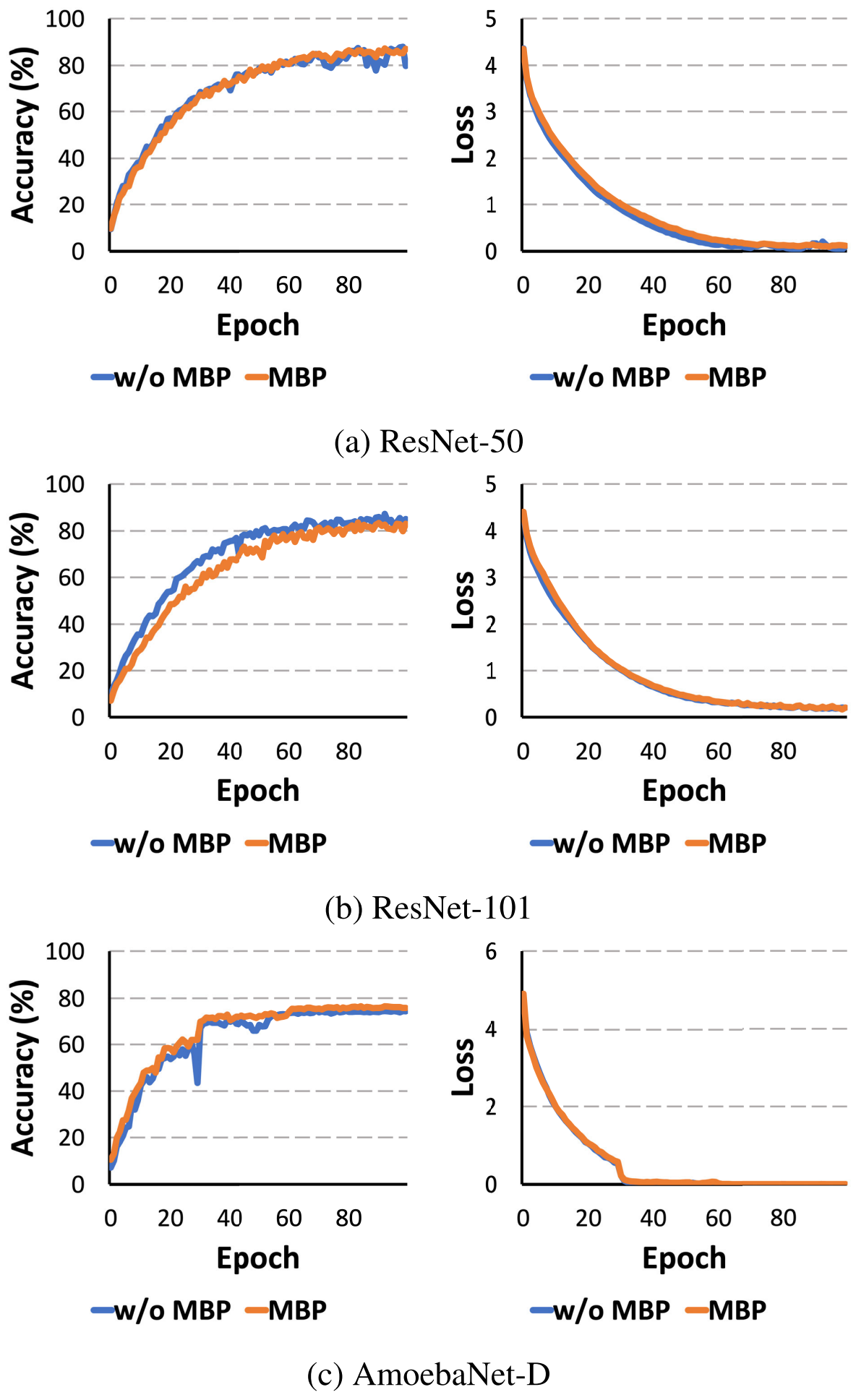

**FIGURE 5. Accuracy and Loss results for three image classification models.**

for MBP is determined to be half the size of mini-batch for comparison.

Figure 5 shows the overall performance of training between models when using and not using MBP for image classification tasks. The loss and accuracy at each epoch are very similar for all methods. This shows that the model using MBP is trained similarly to the model that is not using MBP and this means that the loss normalization method used in this paper is appropriate. This also shows that the MBP method does not affect the model performance during training like other hyper-parameter such as (mini) batch size that is used as the unit of training for various DNN models. Maximum accuracy for each model is shown in Table 4.

**TABLE 3. IoU for semantic segmentation model (U-Net).**

| Metrics | w/o MBP | MBP |
|---|---|---|
| IoU (%) | 95.48 ±0.13 | 95.45 ±0.26 |

To evaluate the results of U-Net, we use a metric called intersection over union (IoU) [21], which is one of metrics to evaluate performance in semantic segmentation tasks and shows the degree of the area of overlap between the predicted image and the ground-truth image. The result in table 3 shows the performance of training between U-Net using MBP and not using MBP. The initial experiment shows that the performance is comparable.

### 2) MAXIMUM BATCH SIZE

If MBP is used, one can compute the entire batch by streaming small batches regardless of the type of models, datasets, entire batch size and the capacity of a device memory, if necessary. Experimental results are shown in table 4 and 5. Three different random seeds are used for the experimental results, and the mean and standard deviation of maximum accuracy/IoU for each random seed are calculated. The size of mini-batch starts from the maximum size that can be computed on the GPU (without using MBP). As the mini-batch sizes of each of the learning models are increased, the model that uses MBP can continue to train while previous methods that does not use MBP fail because of limited memory size.

ResNet-50 model uses 7 different mini-batch sizes, ResNet-101 model uses 8 different mini-batch sizes, and AmoebaNet-D model uses 6 different mini-batch sizes to evaluate MBP performance in this experiment. The U-Net model use 7 different mini-batch sizes to evaluate MBP performance, and the result shows the maximum IoU value for each mini-batch sizes and micro-batch sizes in table 5. The micro-batch size is experimentally determined; 1) the half of mini-batch size in the smallest mini-batch, 2) the maximum size that can compute on GPU in other large mini-batches.

Table 4 and 5 shows the accuracy or IoU for each mini-batch and micro-batch. Without MBP, all models fail to train models using a large mini-batch because the capacity of GPU memory is not enough to allocate a large mini-batch. When using MBP, all models are trained using a large mini-batch up to a given maximum size on the same GPU. Compared to models without MBP, deep learning models are trained using 64× larger mini-batch in ResNet-50 model, 128× larger mini-batch in ResNet-101 model, and 32× larger mini-batch in AmoebaNet-D model. Moreover, the U-Net can train using 64× larger mini-batch sizes compared to the same model without MBP. Although the input data size is 3.45× larger for AmoebaNet-D model and 2.94× larger for U-Net model than the input data size for ResNet models, AmoebaNet-D and U-Net models can increase the size of mini-batch up to the given maximum size when using MBP. Theoretically, MBP allows the increase of the mini-batch size up to the total size of the dataset, if necessary.

**TABLE 4. Accuracy and training time of larger batch sizes than the remaining GPU memory for image classification tasks.**

| Model | Batch size | $\mu$-batch size | Accuracy (%) w/o MBP | Accuracy (%) MBP | Training time (sec) w/o MBP | Training time (sec) MBP |
|---|---|---|---|---|---|---|
| ResNet-50 | 16 | 8 | 87.16 ±0.33 | 87.04 ±1.04 | 226.8 ±0.01 | 233.0 ±0.01 |
| | 32 | 16 | Failed | 86.51 ±2.14 | Failed | 229.4 ±0.02 |
| | 64 | 16 | Failed | 88.96 ±0.41 | Failed | 227.4 ±0.02 |
| | 128 | 16 | Failed | 89.61 ±0.14 | Failed | 227.5 ±0.02 |
| | 256 | 16 | Failed | 88.18 ±0.64 | Failed | 229.4 ±0.02 |
| | 512 | 16 | Failed | 85.95 ±0.11 | Failed | 232.6 ±0.03 |
| | 1024 | 16 | Failed | 77.63 ±0.86 | Failed | 238.2 ±0.02 |
| ResNet-101 | 8 | 4 | 85.82 ±1.84 | 83.54 ±2.51 | 385.5 ±0.02 | 405.0 ±0.01 |
| | 16 | 8 | Failed | 86.35 ±2.19 | Failed | 381.9 ±0.01 |
| | 32 | 8 | Failed | 86.59 ±1.70 | Failed | 381.3 ±0.01 |
| | 64 | 8 | Failed | 85.94 ±1.02 | Failed | 381.5 ±0.01 |
| | 128 | 8 | Failed | 88.14 ±0.42 | Failed | 382.1 ±0.01 |
| | 256 | 8 | Failed | 87.90 ±0.28 | Failed | 383.8 ±0.02 |
| | 512 | 8 | Failed | 82.72 ±0.67 | Failed | 386.4 ±0.03 |
| | 1024 | 8 | Failed | 75.14 ±0.67 | Failed | 392.8 ±0.02 |
| AmoebaNet-D | 32 | 16 | 74.51 ±0.67 | 75.92 ±1.59 | 95.2 ±0.01 | 100.1 ±0.22 |
| | 64 | 32 | Failed | 73.78 ±0.43 | Failed | 95.7 ±0.04 |
| | 128 | 32 | Failed | 73.23 ±1.59 | Failed | 96.4 ±0.02 |
| | 256 | 32 | Failed | 75.80 ±0.17 | Failed | 98.9 ±0.17 |
| | 512 | 32 | Failed | 74.69 ±0.49 | Failed | 103.9 ±0.19 |
| | 1024 | 32 | Failed | 72.49 ±1.10 | Failed | 111.1 ±0.06 |

**TABLE 5. IoU and training time of larger batch sizes than the remaining GPU memory for semantic segmentation task.**

| Model | Batch size | $\mu$-batch size | IoU (%) w/o MBP | IoU (%) MBP | Training time (sec) w/o MBP | Training time (sec) MBP |
|---|---|---|---|---|---|---|
| U-Net | 16 | 8 | 95.48 ±0.13 | 95.45 ±0.26 | 180.7 ±0.18 | 183.2 ±0.20 |
| | 32 | 16 | Failed | 96.13 ±0.13 | Failed | 180.5 ±0.24 |
| | 64 | 16 | Failed | 96.83 ±0.12 | Failed | 182.8 ±0.28 |
| | 128 | 16 | Failed | 97.51 ±0.06 | Failed | 187.0 ±0.24 |
| | 256 | 16 | Failed | 96.96 ±0.16 | Failed | 194.2 ±0.21 |
| | 512 | 16 | Failed | 94.18 ±0.44 | Failed | 207.3 ±0.21 |
| | 1024 | 16 | Failed | 94.18 ±0.44 | Failed | 228.0 ±0.07 |

It is observed that MBP can be used to determine or find the optimal (mini) batch size for each model. Determining the optimal batch size for a deep learning model will affect that model's performance. When using MBP, ResNet-50 model achieved 2.44% higher accuracy in 128 mini-batch sizes, and ResNet-101 model achieved 2.32% higher accuracy in 128 mini-batche sizes and it is shown in table 4. Although AmeobaNet-D model using MBP shows 1.41% higher accuracy in 32 mini-batch size, comparable performance of 1.29% higher is shown in 256 mini-batch size. The U-Net model shows higher percentage of IoU score between 32 and 256 mini-batch sizes as shown in table 5. The U-Net model achieved 97.51% the highest percentage of IoU score in 128 mini-batch size, and it is 2.03% higher compared to the model without MBP. Previously, research on deep learning models had difficulty searching for the optimal batch size and train using a large batch, due to the limitation of hardware resource such as the GPU memory capacity and had to consider multiple GPU(s) and/or virtual memory for GPUs. MBP allows large mini-batch size even on a single memory limited device.

### 3) TRAINING TIME

Deep learning models using MBP process multiple back-propagations for multiple micro-batches until updated for one mini-batch. This is because MBP splits one mini-batch into several micro-batches and streams micro-batches to GPU sequentially. To update the model parameters for one mini-batch, MBP process multiple back-propagations for micro-batches. This process will cause the overhead for training time. Table 4 and 5 show the results for average training time for one epoch between with and without MBP. Deep learning models without MBP show almost less training time compared to models using MBP except for the ResNet-101 model. The results show that average training time will increase as the mini-batch size is increased. This is because the overhead of loading images from secondary memory to CPU memory affects training time more than MBP process overhead. On the other hand, the overhead of training time is relatively small when using the mini-batch size that shows the best performance. The ResNet-50 model shows 1.1% and 0.3% training overhead in 32 and 128 mini-batch size, AmoebaNet-D model shows 5.1% and 4.0% training overhead in 32 and 256 mini-batch size, and U-Net models shows 3.5% training overhead in 128 mini-batch size.

## V. CONCLUSION

This paper introduces Micro-Batch Processing (MBP) that uses a batch streaming method and a loss normalization scheme based on the gradient accumulation to allow deep learning models to train using batches larger than the remaining memory capacity after the model is uploaded, if necessary and while maintaining the performance. Using MBP, this can be done without increasing the memory capacity of the GPU or without using multiple GPUs. In this paper, deep learning models are trained and tested using larger batch sizes previously not attempted because of memory limitation on a single GPU. Results showed overall comparable performance and some tests using larger mini-batch sizes showed higher performance. This shows that although MBP itself is not a method to find the optimal batch size, it can be used to test larger batch sizes and ultimately search for better batch sizes than previously known for a given deep learning model.

MBP will be useful in memory constrained environments using embedded systems. Also, as the learning model becomes larger and data items become larger the memory in any system can be constrained. Therefore, even a faster GPU that has a larger GPU memory size can be constrained with memory and using MBP will relieve that constraint. This means that using MBP, one can train a learning model to use a large mini-batch size up to the whole dataset as a batch in any system. While MBP allows the increase in batch sizes, it cannot enable training if the remaining memory size is too small to accommodate at least one data item. In the future, this limitation may be overcome either by reducing the size of the data item or by allowing partial data item for learning.

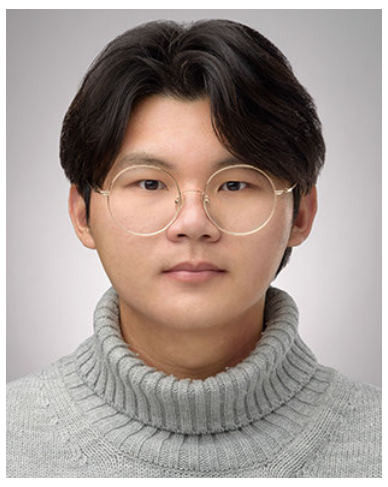

**XINYU PIAO** received the B.S. degree from the School of Electrical Engineering, Korea University, in 2020, where he is currently pursuing the Ph.D. degree at the School of Electrical Engineering. His research interests include deep learning acceleration, system for machine learning, heterogeneous distributed systems, and hardware resource management.

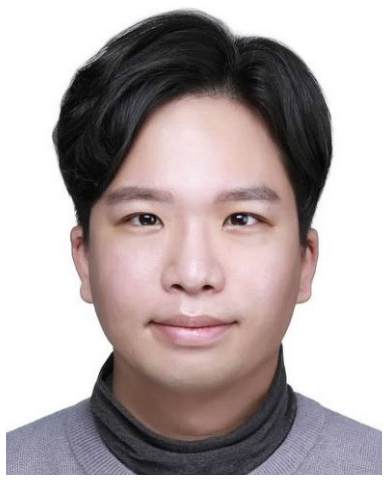

**DOANGJOO SYNN** received the B.S. and M.S. degrees from the School of Electrical Engineering, Korea University, Seoul, in 2020 and 2023, respectively. He is currently pursuing the Ph.D. degree in computer science at Georgia Tech, USA.

From 2012 to 2014, he was a Software Engineer at the KAIST Engineering School. From 2017 to 2020, he was a Software Engineer at the OS and Production Cells, AKA Intelligence Inc., and Flysher Inc. From 2020 to 2021, he was a AWS Cloud Student Ambassador at the Amazon Web Services. From 2020 to 2022, he was a Research Assistant at the High Performance and Intelligent Computing Laboratory, Korea University. From 2021 to 2022, he was a Research Engineer at the Propia AI Inc., Austin, TX, USA. His research interests include distributed computing, machine learning systems, cloud computing, and hyperscalar computing.

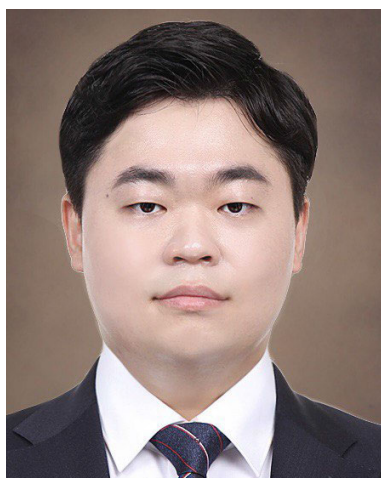

**JOOYOUNG PARK** received the B.S. degree from the School of Electrical Engineering, Korea University, Seoul, South Korea, in 2023. He is currently pursuing the Ph.D. degree at the School of Computer Science and Engineering, Nanyang Technological University, Singapore.

From 2020 to 2022, he was an Undergraduate Student Research Assistant at the High Performance and Intelligent Computing Laboratory, Korea University. His research interests include cloud and serverless computing, heterogeneous distributed computing, and systems for machine learning.

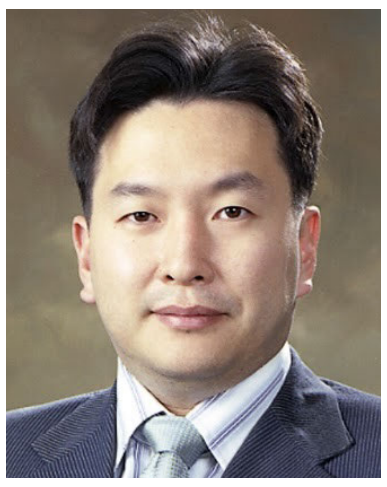

**JONG-KOOK KIM** (Senior Member, IEEE) received the B.S. degree from the Department of Electronic Engineering, Korea University, Seoul, South Korea, in 1998, and the M.S. and Ph.D. degrees from the School of Electrical and Computer Engineering, Purdue University, West Lafayette, USA, in 2000 and 2004, respectively. He was with the Samsung SDS's IT Research and Development Center, from 2005 to 2007. He is currently a Professor at the School of Electrical Engineering, Korea University, where he joined, in 2007. His research interests include heterogeneous distributed computing, energy-aware computing, resource management, evolutionary heuristics, distributed mobile computing, artificial neural networks, efficient deep learning, systems for AI, and distributed robot systems. He is a Senior Member of ACM.